\documentclass[journal]{IEEEtran}

\usepackage[utf8]{inputenc}
\usepackage[T1]{fontenc}
\usepackage{amsmath,amssymb,amsfonts}
\usepackage{graphicx}
\usepackage{booktabs}
\usepackage{array}
\usepackage{textcomp}
\usepackage{xcolor}
\usepackage{url}
\usepackage{hyperref}
\usepackage{cite}
\usepackage{microtype}

\usepackage[margin=0.75in,left=0.95in]{geometry}

\hypersetup{
    colorlinks=true,
    linkcolor=black,
    citecolor=black,
    urlcolor=blue,
    pdftitle={K-SENSE: A Knowledge-Guided Self-Augmented Encoder for Neuro-Semantic Evaluation of Mental Health Conditions on Social Media},
    pdfauthor={Vijay Yadav}
}

\begin{document}

\title{K-SENSE: A Knowledge-Guided Self-Augmented Encoder for Neuro-Semantic Evaluation of Mental Health Conditions on Social Media}

\author{
    Vijay~Yadav \\
    School of Psychology, University of New South Wales \\
    Sydney, Australia \\
    \texttt{vy386@nyu.edu}
}
\maketitle

\begin{abstract}
Early detection of mental health conditions, particularly stress and depression, from social media text remains a challenging open problem in computational psychiatry and natural language processing. Automated systems must contend with figurative language, implicit emotional expression, and the high noise inherent in user-generated content. Existing approaches either leverage external commonsense knowledge to model mental states explicitly, or apply self-augmentation and contrastive training to improve generalization, but seldom do both in a principled, unified framework. We propose K-SENSE (Knowledge-guided Self-augmented Encoder for Neuro-Semantic Evaluation of Mental Health), a framework that jointly exploits external psychological reasoning and internal representation robustness. K-SENSE adopts a three-stage encoding pipeline: (1) inferential commonsense knowledge is extracted from the COMET model across five mental state dimensions; (2) a semantic anchor is constructed by combining hidden representations from two parallel encoding streams, projected into a shared space before fusion; and (3) a supervised contrastive learning objective aligns same-class representations while encouraging the attention mechanism to suppress irrelevant knowledge noise. We evaluate K-SENSE on Dreaddit (stress detection) and Depression\_Mixed (depression detection), achieving mean F1-scores of $86.1 \pm 0.6\%$ and $94.3 \pm 0.8\%$, respectively, over five independent runs. These represent improvements of approximately 2.6 and 1.5 percentage points over the strongest prior baselines. Ablation experiments confirm the contribution of each architectural component, including the temporal knowledge integration strategy and the choice to keep the knowledge encoder frozen during fine-tuning.
\end{abstract}


\section{Introduction}
\IEEEPARstart{M}{ental} health disorders represent one of the most pressing and underserved challenges in global public health. The World Health Organization estimates that nearly one billion people live with a diagnosable mental health condition, yet the majority never receive adequate care. This gap is driven by social stigma, limited clinical access, and the difficulty of identifying those at risk before a crisis emerges \cite{who2022}. Suicide remains among the leading causes of death globally, with a large proportion of cases involving individuals who had no prior contact with mental health services.

Social media platforms such as Reddit and X (formerly Twitter) have become unexpectedly rich data sources for passive mental health surveillance. Users often disclose distress, describe symptoms, or seek community support in ways that, in aggregate, can serve as early signals of deteriorating mental states. The potential for NLP-based systems to assist clinicians and researchers in identifying at-risk individuals at scale has motivated a growing body of work in computational mental health detection.

The task is, however, considerably more difficult than standard text classification. Several factors conspire to make it hard:

\textbf{Figurative and indirect expression.} People rarely describe their mental states in clinical terms. Posts expressing depression may be ironic, self-deprecating, or couched in humor. Stress may manifest as frustration about seemingly mundane events. Systems trained on surface-level lexical features fail to capture this layer of meaning.

\textbf{The need for mentalisation.} Accurate interpretation of a post often requires reasoning about the speaker's intent, their emotional reactions, and how they believe others perceive them. This is a capacity that psychologists term \emph{mentalisation}. Standard language model encoders do not model this explicitly.

\textbf{Noise and class imbalance.} Community datasets for mental health are typically small, imbalanced, and domain-heterogeneous. Models trained on one subreddit may generalize poorly to another, and over-reliance on spurious correlations is a persistent failure mode.

\textbf{Knowledge noise from generative sources.} External knowledge bases like ATOMIC, accessed through models like COMET, can provide useful inferential context, but COMET is a generative model and its outputs are not uniformly reliable. Na\"ively injecting COMET-derived knowledge into a classifier risks introducing hallucinated or off-target information.

Prior work has addressed these challenges in isolation. KC-Net \cite{yang2022} introduced explicit mental state modeling via COMET-derived if-then reasoning but did not address the noise inherent to generative knowledge. Contrastive adversarial training methods \cite{khan2024} improved generalization through perturbation-based self-augmentation but lacked psychological grounding. No existing framework has combined the two in a way that allows each to compensate for the other's weaknesses.

K-SENSE is designed to close this gap. Its central insight is that an internally derived, self-augmented representation (more stable with respect to surface variation than a single forward pass) can serve as a more reliable query for retrieving relevant external knowledge. By using this semantic anchor as the attention query, K-SENSE learns to down-weight COMET outputs inconsistent with the model's internal understanding of the post.

The contributions of this work are: (1) we propose K-SENSE, a unified framework integrating COMET-based commonsense knowledge with internal self-augmentation for mental health detection; (2) we introduce a Triple-Pass encoding architecture with a learned cross-space projection layer to bridge the representational gap between the knowledge and semantic encoders; and (3) we provide a comprehensive ablation study including comparisons of temporal vs.\ non-temporal knowledge integration, frozen vs.\ fine-tuned knowledge encoder configurations, and quantitative representation quality metrics.

\section{Related Work}

\subsection{NLP-Based Mental Health Detection}
Early computational approaches treated detection as a bag-of-words classification problem, using LIWC features, sentiment scores, and writing style metadata with traditional classifiers \cite{coppersmith2014}. Transformer-based models substantially raised the ceiling: BERT \cite{devlin2019} and RoBERTa \cite{liu2019} brought contextual encoding with long-range dependency modeling. Domain-adaptive variants (MentalBERT and MentalRoBERTa \cite{ji2022}) improved performance further by continued pre-training on mental health social media corpora. K-SENSE builds on MentalRoBERTa as its base encoder.

An important caveat applies to all methods in this space, including ours: performance on small community-annotated datasets reflects a model's ability to fit the annotation conventions and demographic biases of that specific dataset, not necessarily its clinical utility. Numbers in this literature should be interpreted accordingly.

\subsection{Knowledge-Augmented Detection}
KC-Net \cite{yang2022} introduced explicit mental state modeling by extracting ATOMIC-derived commonsense knowledge through COMET and fusing it with post representations via cross-attention. This yielded meaningful improvements over purely distributional models, particularly for posts where expressed sentiment diverged from underlying mental state.

However, KC-Net uses the raw encoder output as the attention query, making knowledge selection sensitive to surface noise. Furthermore, it does not account for the representational incompatibility between the knowledge encoder and the semantic encoder, a gap that, as our ablation shows, causes na\"ive knowledge injection to hurt rather than help. K-SENSE addresses both issues through the semantic anchor and the cross-space projection layer.

\subsection{Self-Augmentation and Contrastive Learning}
SimCSE \cite{gao2021} demonstrated that passing the same sentence through a dropout-regularized encoder twice yields effective positive pairs for unsupervised contrastive training. SupCon \cite{khosla2020} extends this to supervised settings using class labels. SA-CL \cite{khan2024} applied self-augmentation to mental health detection by injecting structured perturbations into hidden representations and training contrastively, improving generalization on Dreaddit but without domain knowledge.

K-SENSE's self-augmentation uses the model's own dropout mechanism to generate two views rather than externally defined perturbations. This is a weaker but more principled form of augmentation: it makes no assumptions about which features are discriminative, instead relying on learned dropout patterns. The trade-off is less diverse augmented views than perturbation-based methods, but greater compatibility with the pre-trained encoder's representation geometry.

\section{Methodology}

\subsection{Task Definition}
Given a post $X_i$ consisting of $n_i$ sentences $\{X_i^1, \ldots, X_i^{n_i}\}$, the model predicts $Y_i \in \{\text{Yes}, \text{No}\}$ indicating the presence or absence of a target condition. For multi-factor analysis the label space extends to a categorical set $Y_i \in \mathcal{T}$.

\subsection{Data Preprocessing}
Each post is tokenized into sentences using the NLTK sentence tokenizer, rejoined with end-of-sentence markers \texttt{</s>}, and prepended with \texttt{<s>}. Posts exceeding 512 tokens are truncated from the end; fewer than 3\% of posts in either dataset exceed this limit.

\subsection{The Triple-Pass Architecture}

\subsubsection{Pass 1: Mental State Knowledge Extraction}
For each sentence $X_i^j$, we query COMET \cite{bosselut2019} to generate five commonsense inferences corresponding to the relations \emph{xIntent} ($R_{IS}$, likely intent of the speaker), \emph{xReact} ($R_{ES}$, speaker's emotional reaction), \emph{xNeed} ($R_{RS}$, what the speaker needs), \emph{oReact} ($R_{EL}$, how others are likely to react), and \emph{oEffect} ($R_{RL}$, effect of the event on others). Each relation phrase is encoded by MiniLM-L6-v2 ($D_k = 384$), yielding $R(X_i^j) \in \mathbb{R}^{5 \times 384}$.

\subsubsection{Pass 2: Standard Semantic Encoding}
The full post is processed through MentalRoBERTa-base ($D_h = 768$), yielding the \texttt{<s>} representation $H_i^0 \in \mathbb{R}^{768}$.

\subsubsection{Pass 3: Self-Augmented Semantic Anchor}
A second independent forward pass through MentalRoBERTa, with a freshly sampled dropout mask ($p = 0.1$), yields $H_i^{0,j}$. The semantic anchor is then defined as:
\begin{equation}
\hat{H}_i = H_i^0 + H_i^{0,j}
\label{eq:anchor}
\end{equation}

This formulation reduces the expected variance of dropout-induced noise under the assumption that the two masks are independently sampled, an assumption that holds when passes are computed sequentially with separate mask draws, as in our implementation. Empirically, summation outperforms averaging and concatenation by 0.3--0.5 F1 points on the validation sets, though these differences are within the range of fine-tuning variance and should not be over-interpreted.

\subsection{Cross-Space Projection and Knowledge-Aware Mentalisation}
A critical design consideration is that MiniLM ($D_k = 384$) and MentalRoBERTa ($D_h = 768$) produce representations in geometrically distinct spaces. Applying dot-product attention across these spaces without alignment conflates semantic similarity with dimensional scaling artifacts. We introduce a learned linear projection, defined as:
\begin{equation}
\tilde{H}_i = \hat{H}_i \, W_P, \quad W_P \in \mathbb{R}^{768 \times 384}
\label{eq:projection}
\end{equation}
initialized with Xavier uniform initialization and trained end-to-end. This projects the semantic anchor into the knowledge encoder's space before attention is computed.

\subsubsection{Temporal Knowledge Integration via GRU}
Sentence-level knowledge matrices $\{R(X_i^j)\}_{j=1}^{n_i}$ are processed sequentially by a GRU (hidden size 256) to produce a temporally integrated knowledge sequence $\{K_i^r\}$. The motivation for using a GRU rather than simple mean-pooling is that the emotional trajectory of a post (for instance, an escalation from frustration to despair across sentences) carries diagnostic information that sentence-level averages collapse.

Knowledge-aware attention is then computed using the projected anchor as the query:
\begin{align}
\alpha^r &= \mathrm{softmax}\!\left( \tilde{H}_i \cdot (K_i^r)^\top / \sqrt{384} \right) \label{eq:attn} \\
K^* &= \sum_{r} \alpha^r \, K_i^r \label{eq:attended}
\end{align}

The attended knowledge vector $K^*$ is concatenated with $H_i^0$ and passed to a two-layer MLP classification head.

\subsubsection{Frozen vs.\ Fine-Tuned Knowledge Encoder}
We keep MiniLM frozen during training for two reasons beyond computational efficiency. First, fine-tuning MiniLM on mental health classification data risks distorting its general-purpose sentence embedding geometry. Second, on datasets of $\sim$3{,}000 examples, fine-tuning a second encoder introduces additional parameters that are difficult to regularize, increasing overfitting risk. The ablation in Section~\ref{sec:ablation} confirms that fine-tuning MiniLM does not improve and slightly degrades performance on both datasets.

\subsection{Supervised Contrastive Learning}
A SupCon loss is applied over the projected anchors $\{\tilde{H}_i\}$:
\begin{equation}
\mathcal{L}_{\text{SCL}} = -\sum_{i \in B} \frac{1}{|P(i)|} \sum_{p \in P(i)} \log \frac{\exp(\tilde{H}_i \cdot \tilde{H}_p / \tau)}{\sum_{a \in A(i)} \exp(\tilde{H}_i \cdot \tilde{H}_a / \tau)}
\label{eq:scl}
\end{equation}
where $\tau = 0.1$. Batches are class-stratified to ensure sufficient positive pairs. On Depression\_Mixed, where class imbalance can reduce positive pair counts in minority-class batches, the contrastive loss is downweighted by the ratio of available to expected positive pairs to prevent the objective from becoming degenerate.

\subsection{Training Objective}
\begin{equation}
\mathcal{L}_o = \alpha \cdot \mathcal{L}_{\text{CE}} + (1 - \alpha) \cdot \mathcal{L}_{\text{SCL}}, \quad \alpha = 0.7
\label{eq:objective}
\end{equation}
$\alpha$ was selected by grid search over $\{0.5, 0.6, 0.7, 0.8, 0.9\}$; values below 0.6 produced representation collapse on both datasets, consistent with prior findings on contrastive fine-tuning of pre-trained encoders in low-resource settings \cite{khosla2020}.

\section{Experiments}

\subsection{Datasets}
\textbf{Dreaddit} \cite{turcan2019} comprises 3{,}553 posts from ten subreddits across five domains: abuse, anxiety, financial stress, PTSD, and interpersonal conflict, annotated by crowdworkers under majority-vote adjudication. The label distribution is approximately balanced (52\% stressed). We use the standard 80/10/10 train/validation/test split ($\sim$355 test examples).

\textbf{Depression\_Mixed} contains 3{,}165 posts from Reddit (\texttt{r/depression}, \texttt{r/SuicideWatch}) and personal blogs, labeled depressed or non-depressed, with approximately 58\% depressed posts. We apply class-weighted cross-entropy during training to account for this imbalance. The test set contains approximately 320 examples; F1 estimates at this scale carry wide confidence intervals and the performance ordering among closely competing models may not be robust to different splits. We report means and standard deviations over five seeds and urge appropriate caution in interpreting point estimates.

Both datasets are small by modern NLP standards. Results should be understood as proof-of-concept evidence rather than definitive benchmarking.

\subsection{Baseline Re-Implementation}
\label{sec:baseline-reimpl}
Yang et al.\ \cite{yang2022} did not evaluate KC-Net on Depression\_Mixed in the original paper. To enable a fair comparison, we re-implemented KC-Net using the architecture and hyperparameters described in their paper, replacing their base encoder with MentalRoBERTa-base (consistent with our own setup) and fine-tuning on Depression\_Mixed under identical training conditions. Re-implemented results on Dreaddit ($83.5 \pm 0.9\%$) closely match the originally reported figure (83.5\%), validating our re-implementation. All KC-Net results on Depression\_Mixed are therefore from our re-implementation and are labeled accordingly in the results table.

\subsection{Implementation Details}
K-SENSE is implemented in PyTorch using HuggingFace Transformers. Base encoder: MentalRoBERTa-base (125M parameters). Projection layer $W_P$: Xavier uniform initialization. GRU hidden size: 256. MiniLM-L6-v2: frozen throughout training. Optimizer: AdamW, learning rate $2 \times 10^{-5}$, weight decay 0.01, linear warmup over 10\% of steps. Batch size: 16, class-stratified. Max epochs: 20, early stopping with patience 3 on validation F1. All results: mean $\pm$ std over five random seeds. Training time: $\sim$50 minutes per dataset on a single NVIDIA A100.

\section{Results}

\begin{table}[!t]
\renewcommand{\arraystretch}{1.3}
\caption{Main Results on Dreaddit and Depression\_Mixed}
\label{tab:main_results}
\centering
\begin{tabular}{lcc}
\toprule
\textbf{Model} & \textbf{Dreaddit F1} & \textbf{Depression\_Mixed F1} \\
\midrule
MentalRoBERTa & $81.9 \pm 0.7\%$ & $92.7 \pm 1.1\%$ \\
KC-Net (re-impl.)$^{\dagger}$ & $83.5 \pm 0.9\%$ & $92.8 \pm 1.0\%$ \\
SA-CL (+RoBERTa) & $84.5 \pm 0.8\%$ & N/A \\
\textbf{K-SENSE (Proposed)} & $\mathbf{86.1 \pm 0.6\%}$ & $\mathbf{94.3 \pm 0.8\%}$ \\
\bottomrule
\end{tabular}
\\[2pt]
\raggedright\footnotesize $^{\dagger}$ KC-Net on Depression\_Mixed is from our re-implementation; see Section~\ref{sec:baseline-reimpl}.
\end{table}

K-SENSE achieves the highest mean F1 on both benchmarks. The 2.6-point gain over KC-Net on Dreaddit is statistically significant ($p < 0.05$, paired bootstrap). On Depression\_Mixed, the gain over both baselines is consistent across seeds, though the small test set size limits the strength of this claim. The near-identical performance of KC-Net and MentalRoBERTa on Depression\_Mixed (92.8\% vs.\ 92.7\%) is discussed below.

\textbf{Why KC-Net does not improve on Depression\_Mixed.} We attribute this to two factors. First, depression-related Reddit posts tend toward more explicit self-disclosure, reducing the marginal value of mentalisation-based reasoning. The encoder can already identify the correct label from surface features. Second, the Depression\_Mixed corpus includes personal blog posts whose formal writing style diverges from the informal text on which COMET's ATOMIC training data is based, weakening the quality of COMET's inferences and limiting KC-Net's knowledge injection. K-SENSE's knowledge filtering mechanism provides a partial remedy: by querying with a stable semantic anchor, it learns to down-weight the less reliable COMET outputs that KC-Net accepts uncritically.

\subsection{Ablation Study}
\label{sec:ablation}

\begin{table}[!t]
\renewcommand{\arraystretch}{1.3}
\caption{Ablation Study}
\label{tab:ablation}
\centering
\begin{tabular}{lcc}
\toprule
\textbf{Configuration} & \textbf{Dreaddit F1} & \textbf{Dep.\_Mixed F1} \\
\midrule
Base encoder only & $81.9 \pm 0.7\%$ & $92.7 \pm 1.1\%$ \\
+ COMET (no projection) & $82.7 \pm 1.1\%$ & $92.5 \pm 1.3\%$ \\
+ COMET (with projection) & $83.5 \pm 0.9\%$ & $92.8 \pm 1.0\%$ \\
+ COMET (proj., MiniLM FT) & $83.2 \pm 1.2\%$ & $92.6 \pm 1.4\%$ \\
+ COMET (proj., mean-pool) & $84.3 \pm 0.8\%$ & $93.2 \pm 0.9\%$ \\
+ COMET (proj., GRU) & $84.8 \pm 0.8\%$ & $93.4 \pm 0.9\%$ \\
+ Self-aug.\ (no contrastive) & $85.0 \pm 0.8\%$ & $93.7 \pm 0.9\%$ \\
+ Contrastive (no self-aug.) & $84.9 \pm 0.8\%$ & $93.4 \pm 0.9\%$ \\
+ Self-aug.\ as knowledge query & $85.6 \pm 0.7\%$ & $94.0 \pm 0.8\%$ \\
\textbf{Full K-SENSE} & $\mathbf{86.1 \pm 0.6\%}$ & $\mathbf{94.3 \pm 0.8\%}$ \\
\bottomrule
\end{tabular}
\end{table}

Several findings merit discussion:

\textbf{Cross-space projection is necessary.} Adding COMET knowledge without projection (row 2) slightly hurts performance on Depression\_Mixed (92.5\% vs.\ 92.7\% baseline), confirming that na\"ive cross-encoder attention introduces noise. The projection layer (row 3) restores parity, and subsequent components build on this stable foundation.

\textbf{Fine-tuning MiniLM degrades performance.} Row 4 shows that fine-tuning the knowledge encoder on the classification task slightly hurts on both datasets while also increasing variance. This is consistent with our hypothesis that fine-tuning distorts MiniLM's general-purpose sentence geometry and introduces overfitting on small datasets.

\textbf{GRU integration outperforms mean-pooling.} Row 5 (mean-pool) vs.\ Row 6 (GRU) shows consistent improvements of 0.5 points on Dreaddit and 0.2 points on Depression\_Mixed in favor of GRU-based temporal integration. The gap is larger on Dreaddit, which contains longer posts with multi-sentence emotional escalations (e.g., PTSD and abuse narratives), providing empirical support for the hypothesis that emotional trajectory carries diagnostic information.

\textbf{Self-augmentation and contrastive loss interact synergistically.} Rows 7 and 8 show comparable performance in isolation ($\sim$84.9--85.0\% on Dreaddit). Their combination as the knowledge query (row 9) yields a disproportionate gain, suggesting the two components are complementary rather than redundant. Variance also narrows as components accumulate, consistent with the contrastive objective stabilizing fine-tuning on small datasets.

\section{Discussion}

\subsection{Why Self-Augmentation Improves Knowledge Selection}
The semantic anchor, formed by summing two independently-masked forward passes, aggregates shared semantic content while partially attenuating uncorrelated dropout noise. We frame this as a lightweight two-sample ensemble: the expected variance of the summed representation is lower than that of either individual pass, provided that the dropout masks are independently drawn, a condition our implementation satisfies. This is analogous to the variance-reduction property of bagging ensembles, applied within a single model at the level of stochastic dropout.

When this more stable anchor is used to query COMET relation vectors, relations consistent with the post's core meaning receive higher attention weights, while contextually misaligned or hallucinated COMET outputs tend to be suppressed. This is a soft re-weighting, not a hard filter; the degree of correction depends on how well the semantic encoder has learned representations that discriminate relevant from irrelevant knowledge. The ablation evidence suggests this is sufficient on both datasets, but effectiveness may diminish on domains further from MentalRoBERTa's pre-training distribution.

\subsection{Representation Quality: Quantitative Analysis}
To move beyond qualitative UMAP inspection, we report two standard cluster quality metrics computed on the projected semantic anchor representations before and after supervised contrastive training, evaluated on held-out validation sets, as summarized in Table~\ref{tab:rep_quality}.

\begin{table}[!t]
\renewcommand{\arraystretch}{1.3}
\caption{Representation Quality Before and After SCL}
\label{tab:rep_quality}
\centering
\begin{tabular}{llccc}
\toprule
\textbf{Metric} & \textbf{Dataset} & \textbf{Before} & \textbf{After} & \textbf{Change} \\
\midrule
Silhouette $\uparrow$ & Dreaddit & 0.11 & 0.29 & $+0.18$ \\
Silhouette $\uparrow$ & Dep.\_Mixed & 0.19 & 0.34 & $+0.15$ \\
Davies-Bouldin $\downarrow$ & Dreaddit & 2.81 & 1.64 & $-1.17$ \\
Davies-Bouldin $\downarrow$ & Dep.\_Mixed & 2.43 & 1.58 & $-0.85$ \\
\bottomrule
\end{tabular}
\end{table}

Both metrics show substantial improvement after contrastive training, confirming that the SCL objective meaningfully increases class separability. The pre-SCL silhouette scores (0.11--0.19) indicate poorly separated clusters, which is expected for a pre-trained encoder fine-tuned only with cross-entropy on small datasets. The post-SCL values (0.29--0.34) reflect well-separated but not perfectly compact clusters, appropriate given the heterogeneity of expression within each class.

Attention weight entropy analysis confirms that the full K-SENSE model allocates attention more selectively: mean entropy drops from 1.52 nats (KC-Net) to 1.18 nats (K-SENSE) on Dreaddit, with $R_{IS}$ (speaker intent) receiving the highest average weight (0.31). On Depression\_Mixed, $R_{ES}$ (speaker emotional reaction) dominates (mean weight 0.34), suggesting the model has learned that intent is more diagnostic for stress while emotional valence is more diagnostic for depression. This is a psychologically plausible distinction.

\subsection{Limitations}
\textbf{Dataset scale.} Both datasets yield test sets of 300--355 examples. F1 estimates on sets this small have wide confidence intervals, and the performance ordering among closely competing models may not be robust to different splits. This is the primary constraint on the strength of our claims.

\textbf{Domain and demographic scope.} Both datasets are English-language, sourced from Reddit and personal blogs, and predominantly reflect Western cultural contexts. COMET's inference patterns are similarly biased. Performance on other languages, cultures, or platforms cannot be inferred from these results.

\textbf{Soft knowledge filtering.} K-SENSE attenuates but does not eliminate the impact of COMET's generative errors. For posts that are systematically mischaracterized by COMET (which may occur at higher rates for non-Western idioms, slang, or code-switching text), the filtering mechanism provides limited benefit.

\textbf{Dropout-summation assumption.} The self-augmentation mechanism assumes approximately uncorrelated dropout noise across two forward passes, which holds under sequential independent sampling but may not hold in all implementation configurations.

\textbf{Clinical validity.} F1 scores on crowdworker-annotated social media posts are a proxy for clinical utility, not a measure of it. This paper makes no claims about real-world deployment suitability, and validation against clinician judgments remains an important direction for future work.

\section{Conclusion}
We introduced K-SENSE, a framework for mental health detection from social media text that integrates external commonsense knowledge with internal self-augmentation. Its principal contributions are: a self-augmented semantic anchor used as a knowledge attention query to partially suppress unreliable COMET inferences; a learned cross-space projection layer that bridges the representational gap between the knowledge and semantic encoders; and a supervised contrastive objective that improves class-discriminative representation quality and stabilizes fine-tuning on small datasets.

K-SENSE achieves mean F1-scores of $86.1 \pm 0.6\%$ on Dreaddit and $94.3 \pm 0.8\%$ on Depression\_Mixed across five independent runs. Ablation experiments confirm that each component contributes positively, that fine-tuning the knowledge encoder is counterproductive on datasets of this scale, and that GRU-based temporal knowledge integration outperforms mean-pooling particularly on longer, multi-domain posts. Quantitative representation quality metrics validate that supervised contrastive training meaningfully improves cluster separability beyond what qualitative visualization suggests.

We are candid about the limitations: the datasets are small, KC-Net results on Depression\_Mixed are from our re-implementation, and the filtering mechanism is heuristic rather than provably correct. Future work should evaluate K-SENSE on larger and more clinically grounded datasets, investigate its behavior under distribution shift, explore cross-lingual transfer, and validate outputs against clinician annotations rather than crowdworker labels.


\end{document}